\documentclass[preprint]{elsarticle}

\usepackage{graphicx}
\usepackage{amsmath,amssymb,amsfonts}

\usepackage{lineno}
\usepackage{textcomp}
\usepackage{empheq} 
\usepackage{xcolor}
\usepackage{optidef}
\usepackage{subcaption}
\usepackage{colortbl}
\usepackage[algo2e,linesnumbered,ruled]{algorithm2e} 
\usepackage{algorithm}
\usepackage{xspace}
\usepackage{multirow}

\newcommand{\modelname}{\lambda}
\newcommand{\model}{{\tt CorrDetector}\xspace}

\biboptions{sort&compress}

\journal{Expert Systems with Applications}

\begin{document}

\begin{frontmatter}



\title{CorrDetector: A Framework for Structural Corrosion Detection from Drone Images using Ensemble Deep Learning}





\author[sut]{Abdur Rahim Mohammad Forkan\corref{cor1}}
\ead{fforkan@swin.edu.au}
\author[sut2]{Yong-Bin Kang}
\ead {ykang@swin.edu.au}
\author[sut]{Prem Prakash Jayaraman}
\ead {pjayaraman@swin.edu.au}
\author[sut,acu]{Kewen Liao}
\ead {Kewen.Liao@acu.edu.au}
\author[sut]{Rohit Kaul}
\ead {rkaul@swin.edu.au}
\author[ncastle]{Graham Morgan}
\ead {graham.morgan@newcastle.ac.uk}
\author[ncastle]{Rajiv Ranjan}
\ead {raj.ranjan@ncl.ac.uk}
\author[rai]{Samir Sinha}
\ead {samir@robonomics.ai}

\cortext[cor1]{Corresponding Author}
\address[sut] {Department of Computer Science and Software Engineering, Swinburne University of Technology, Melbourne, Victoria, Australia}
\address[sut2] {Department of Media and Communication, Swinburne University of Technology, Melbourne, Victoria, Australia}
\address[acu] {Discipline of Information Technology, Peter Faber Business School, Australian Catholic University, Sydney, NSW, Australia}
\address[ncastle] {Newcastle University, Newcastle upon Tyne, NE1 7RU, UK} 
\address[rai] {Robonomics AI, York Street, Sydney, NSW, Australia}


\begin{abstract}

In this paper, we propose a new technique that applies automated image analysis in the area of structural corrosion monitoring and demonstrate improved efficacy compared to existing approaches. Structural corrosion monitoring is the initial step of the risk-based maintenance philosophy and depends on an engineer's assessment regarding the risk of building failure balanced against the fiscal cost of maintenance. This introduces the opportunity for human error which is further complicated when restricted to assessment using drone captured images for those areas not reachable by humans due to many background noises. The importance of this problem has promoted an active research community aiming to support the engineer through the use of artificial intelligence (AI) image analysis for corrosion detection. In this paper, we advance this area of research with the development of a framework, \model. \model uses a novel ensemble deep learning approach underpinned by convolutional neural networks (CNNs) for structural identification and corrosion feature extraction. We provide an empirical evaluation using real-world images of a complicated structure (e.g. telecommunication tower) captured by drones, a typical scenario for engineers. Our study demonstrates that the ensemble approach of \model significantly outperforms the state-of-the-art in terms of classification accuracy.
\end{abstract}

\begin{keyword}
Corrosion detection \sep Object detection \sep Deep Learning \sep Drone Images \sep Industrial structure \sep Ensemble model \sep CNN.


\end{keyword}

\end{frontmatter}


\section{Introduction}
\label{sec:intro}

Inspecting faults (e.g. corrosion) is a major problem in industrial structures such as building roofs, pipes, poles, bridges, and telecommunication towers \citep{fang2020novel}. This is a vital service for several industrial sectors, especially manufacturing, where structures (assets) that are subject to corrosion due to their exposure to the weather are used to deliver critical products or services. The problem of corrosion may cost Australia up to \$32 billion annually, which is greater than \$1500 for every Australian each year \citep{javaherdashti2017microbiologically}. Corrosion is not simply a financial cost if left unattended; the endangerment of lives may also be a real risk. Without adopting to the latest in AI-driven solutions, businesses are losing millions in time and money to identify corrosion using methods that have changed little with heavy reliance on human judgement \citep{javaherdashti2000corrosion}. The timely and accurate detection of corrosion is a key way to improve the efficiency of economy by instigating appropriately managed maintenance processes that will also safe lives.

A fast and reliable inspection process for corrosion can ensure industrial assets are maintained in time to prevent regulatory breaches, outages or catastrophic disasters. In most cases, inspections of such assets are conducted manually which can be slow, hazardous, expensive and inaccurate. Recently, drones have proven to be a viable and safer solution to perform such inspections in many adverse conditions by flying up-close to the structures and take a very large number of high-resolution images from multiple angles \citep{petricca2016corrosion}. The images acquired through such process are stored and then subsequently reviewed manually by expert engineers who decide about further actions. However, this causes a problem of plenty for highly qualified engineers to manually identify corrosion from the images which further leads to a high level of human error, inconsistencies, high lead time and high costs in terms of man-hours. 

Existing approaches for identifying structural corrosion from images are either based on Computer Vision (CV) \citep{fernandez2013automated} or Deep Learning (DL) techniques \citep{cha2018autonomous,atha2018evaluation}. In recent CV-based techniques \citep{jahanshahi2013effect}, non-trivial prior knowledge and extensive human efforts are required in designing high quality corrosion features from images. In addition, one cannot hope much on the performance (or the accuracy of corrosion detection) in the case that the corrosion features are somewhat incorrectly identified. Compared with computer vision/image processing \citep{acosta2014innovative} and vanilla machine learning approaches \citep{son2014rapid},  DL-based methods, in particular Convolutional Neural Networks (CNNs) \citep{gu2018recent,jayaraman2020healthcare} have shown the ability to automatically learn important features, outperforming state-of-the-art vision-based approaches \citep{atha2018evaluation, cha2018autonomous, hoskere2018vision} and achieving human-level accuracy.

In this paper, we present a Deep Learning (DL)-based framework named \model, for detecting corrosion from high resolution images captured by drones.  As the key innovation, we propose and develop an ensemble of CNN models \citep{zheng2020ensemble} which is capable of detecting corrosion in target structure (i.e. object) from such high resolution images at significantly higher accuracy than the current state-of-the-art CNN models. More specifically, the proposed framework is capable of providing i) industrial structure recognition - detect  the industrial structure (i.e. object of interest) in the image captured by the drone (since the drone image is captured in a real-world environment that is filled with background noise) and; iii) localised detection of corrosion - detect which areas in the industrial structure contains corrosion. Most DL-based solutions for corrosion detection use image samples captured by DSLR (digital single-lens reflex), digital or mobile cameras with human involvement in taking pictures \citep{atha2018evaluation, cha2017deep} in more controlled environment. Such image samples are much lower in resolution than drone images. Moreover, these samples can be biased as they are captured specifically to be utilised for experimental purposes at certain distances and angles. Therefore, such images comprised of human judgements to focus in specific type of corrosion area within the image which can be easily isolated and distinguishable even in visual inspection \citep{atha2018evaluation}. Moreover, previous studies have mostly focused on corrosion identification only in metallic surfaces  \citep{son2014rapid, petricca2016corrosion, atha2018evaluation, cha2018autonomous}. To the best of our knowledge, this work is the first attempt that utilises real-world high-resolution unaltered images captured by drones in industrial and real-world settings to identify corrosion in industrial structure such as telecommunication tower. More specifically, this paper makes the following contributions:

\begin{itemize}
    \item Present a novel framework, \model with a 4-layer architecture to detect industrial object and identify regions of corrosion in high-resolution images of industrial assets captured by drones in a real-world setting from various positions, angles and distances. 
    
    \item Present an innovative ensemble approach that combines two deep learning models; a deep learning model for recognising and separating targeted industrial structure from the background and a deep learning model to identify corrosion in specific regions of the industrial structure (localised).   
    
    \item Present a systematic methodology for training our ensemble model using high-resolution drone images that includes two types of annotation techniques namely grid-based and object-based.
    
    
    \item A comprehensive evaluation using a real-world dataset (high resolution drone images of telecommunication towers) and comparison with current state-of-the-art deep learning models for corrosion detection to demonstrate the efficacy of the proposed \model. 
    
    
\end{itemize}

The rest of the paper is organised as follows. Section \ref{sec:literature} provides a discussion of current state-of-the-art in corrosion detection from images. Section \ref{sec:method} presents the systematic methodology for developing an ensemble of CNN models for corrosion detection. Section  \ref{sec:evaluation} presents the experimental domain, the empirical evaluation and a comprehensive analysis of our proposed approach against the current-state-of-the-art and finally Section \ref{sec:conclusion} concludes the paper.

\section{Related Works}
\label{sec:literature}

The most recent research identify the possibility of the utilisation of computer vision (CV) \citep{hanzaei2017automatic} coupled with deep learning (DL) \citep{liu2020survey} for defect identification. This has been proven in the literature to be capable of identifying corrosion in building structures. As this may lower the opportunity of human error while greatly speeding up the analysis process there has been a significant increase in active research pursuing approaches based on these two cornerstones. 

Several CV-based approaches have been proposed for identifying defects in industrial structures to aid in civil engineering maintenance life-cycle service. In particular, structures prone to corrosion through repeated erosion may result in loss of life (e.g., flight, at-sea) and are particular drivers for such work.

Authors in \citep{Taha:2006, fernandez2013automated} used wavelet transforms on images to detect structural corrosion. Assuming images are obtained through nondestructive imaging (NDI) of ageing aircraft materials and structures, the work in \citep{Taha:2006} attempts to identify damage on ageing aircraft structures. Wavelet analysis was used for feature extraction, a clustering technique is used for damage segmentation, and a K-means distance-based method is used for damage classification. \citep{fernandez2013automated} used wavelet features and the Shannon entropy method for detecting damages in ship hulls.

The work in \citep{Ramana:2007} focuses on corrosion surface damage identification in the form of pitting and micro-cracks in metal using an image analysis based on wavelet transforms. The work \citep{Vikram:2010} uses camera based image analysis techniques to identify, quantify and classify damage in aluminium structures. The proposed techniques used the optical contrast of the corroded region with respect to its surroundings, performed edge detection techniques through image processing approaches and computed each region to predict the total area of the affected part. Authors in \citep{son2014rapid} adopted color space features and J48 decision tree classification for detecting rust in steel bridges. They obtained high accuracy (97.51\%) but only used a total of 165 images. Overall, in all CV-based approaches the focus is on feature extraction techniques rather than classification whereas in a DL-based solution using convolutional neural networks (CNNs) there is an opportunity for automatic feature extraction that may lead to more optimum and adaptable learning strategies. This provides an opportunity for a broader area of subject matter to be considered by a single approach (coupled with transferred learning \citep{gu2018recent}) as corrosion types are different as can be seen in the related works just described. 


Within the domain of inspecting industrial structures, Authors in \citep{cha2018autonomous} adopt `Faster R-CNN' for detecting cracks in metal objects. Their database contains 2366 images (with 500 $\times$ 375 pixels) labelled with five types of damages - concrete crack, steel corrosion  classed as medium or high, bolt corrosion, and steel delamination. To develop a training set, 297 images (with a resolution of 6,000 $\times$ 4,000 pixels) were collected using a Nikon D5200 and D7200 DSLR cameras. Images were taken under different lighting conditions. Their evaluation results showed 90.6\%, 83.4\%, 82.1\%, 98.1\%, and 84.7\% average precision (AP) ratings for the five damage types respectively with a mean AP of 87.8\%. The traditional CNN based method showed high accuracy in this and their  previous work \citep{cha2017deep}. 

For the task of corrosion detection, the work in \citep{atha2018evaluation} applied two CNN networks (ZF Net and VGG). However, the authors focused on the computational aspects by proposing two shallow CNN architectures `Corrosion7' and `Corrosion5'. These two networks were shown to be similar to ZF and VGG in terms of detection performance. Taking a closer look at the work of \citep{atha2018evaluation}, during its training phase, authors used color spaces (RGB, YCbCr, CbCr, and grayscale). After determining the optimal color space, they identified that CbCr to be the most robust for corrosion detection using wavelet decomposition. Using the chosen color space, CbCr, authors proposed CNN architectures with a sliding window using different sizes (i.e. 32x32, 64x64, and 128x128). This sliding window approach was used to detect corroded areas within an image. The authors achieved 96.68\% mean precision using 926 images. However, they only focus on clearly visible surface images whereas, our image samples are from a complicated industrial structure. Other corrosion detection model using deep learning did not obtain accuracy over 88\% using the test dataset of targeted industrial assets \citep{petricca2016corrosion, hoskere2018vision}.

The approaches described in the literature rely on CNN approaches that could be advanced by bringing together techniques. Indeed, the transfer learning of the CNN is already established as a technique with learning images constructed efficiently through the transfer of knowledge \citep{gu2018recent}. We considered the possibility of an ensemble technique, using the latest CNN approaches that provide pixel level masking for object identification (Mask R-CNN \citep{he2017mask}) together with localised corrosion detection. Such ensemble approach for corrosion detection differs from existing work in the literature since the image samples are captured in a controlled setting (i.e. images are captured for the purpose of empirical evaluations). On contrary, our proposed approach works on unaltered drone images collected in a real-world setting (via a drone used by engineers to manually assess structural corrosion) which presents high level of complexity (such as  filled with background noise, includes complex industrial structure and captured at different angles, distance and different times of the day i.e overcast, sunny etc.).


\section{The \model Framework}
\label{sec:method}

\begin{figure}[!htb]
    \centering
    \includegraphics[width=4in]{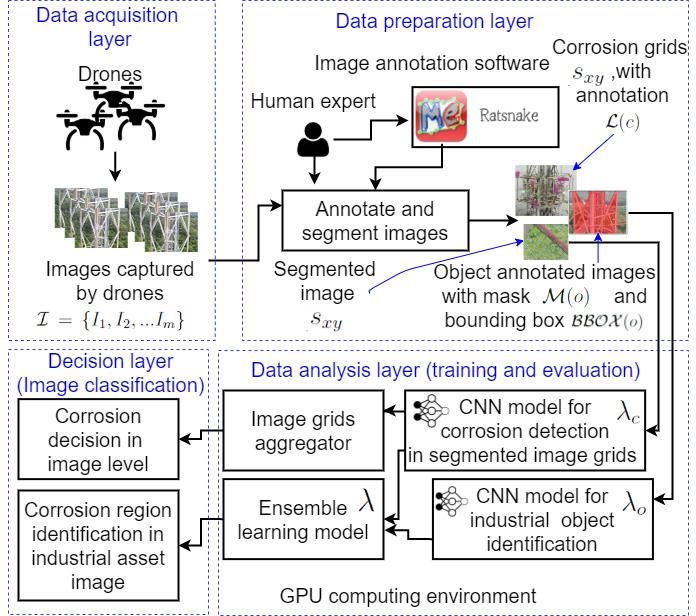}
    \caption{The architecture of the \model framework}
    \label{fig:architecture}
\end{figure}

The architecture of \model framework is illustrated in Figure \ref{fig:architecture}. In Data acquisition layer, a large number of images of the targeted industrial structure (e.g. building roofs, telecommunication towers, bridges, poles, wires and pipelines) are acquired through human-operated drones with advanced cameras. Using these images, engineers identify areas of defects (e.g. corrosion) by visual inspection. The human expert annotated images using various annotation software tools are utilised in data preparation layer.

\model incorporates two types of annotations. Given an image, the first type is the \textit{grid-based annotation} where the multiple small rectangular grids (or segments)~\footnote{In this paper, to simplify the presentation, we interchangeably use the terms, grid and segment.} in the image are drawn. Then, a human expert annotates each of these grids with corrosion or non-corrosion, depending on the fact that the grid contains corroded structural component(s). The grids are separated using image processing techniques. These annotations are used to build ground truth information for corrosion detection deep learning model in \model. Another type of the annotation is the \textit{object annotation}, where a human expert uses an image annotation tool to create a polygonal mask annotation around the target object (e.g. a telecommunication tower on which corrosion needs to be detected). The annotated data are used to train an object recognition model in \model. 

In data analysis layer, two deep learning models using different CNNs are learned to develop an ensemble model in \model. These models are denoted as $\modelname_c$ and $\modelname_o$ that will be used for grid-based corrosion detection and industrial object identification, respectively  The \textit{image grids aggregator} is used to combine all segments (or grids) within the target image to predict whether there is corrosion or not in the image. As another key model, \model also incorporates, an ensemble model, denoted as $\modelname$, that takes the outputs predicted by the $\modelname_c$ and $\modelname_o$ as the input, and predict which structural objects has corrosion or not. This ensemble approach aims to improve our prediction capability by predicting corrosion on only the structural components.

The decision layer makes the final two decisions for a given unknown image. The first decision is the outcome that contains corrosion or not in the image. The decision is made by aggregating outcomes of individual segments estimated using $\lambda_c$ for that image. The other decision is about which regions of industrial object are likely to have corrosion in the image. This is done using the ensemble model, $\lambda$. The corrosion regions can be visualised as highlighted rectangular grids.


\subsection{Image annotation methodology}
In this section we present the image annotation methodology employed by \model in the \textit{data preparation layer}. 

Let $\mathcal{I} = \{I_1, I_2,...I_m\}$ be list of $m$ images captured by drones in the {data acquisition layer}. Each image $I_{i \in [1,m]}$, has a $W \times H$ resolution. 

In our work, given an image, image annotation means annotating the image with a caption that best explains the image based on the prominent objects present in that image and to make the objects recognisable for machines. 
The annotation is done by humans manually using image annotation tools to create the ground truth data for $\model$. An image having places of corrosion is annotated with a cropped bounding box so that other parts of the image can be considered as with no corrosion. As described previously, we apply two types of annotation, grid-based annotation and object annotation on each image, $I_i$. 


\subsubsection{Grid-based image annotation}

The objective of the grid-based corrosion detector, $\lambda_c$ is to predict whether a region in a given image with has corrosion or not. Thus, we formulate the prediction problem for $\lambda_c$ as a binary (1/0) classification, where 1 means corrosion and 0 otherwise. For this kind of classification, the \textit{image segmentation} approach has been proven useful~\citep{dagli2004framework}. In this approach, a given original image (in our case, each $I_i \in \mathcal{I}$), it is segmented equally to a number of rectangular grids. Then, each of the grids is annotated with 1 meaning corrosion, and 0 meaning non-corrosion. This is also a kind of data upsampling or augmentation approach to create enough data to train CNN models.

In our approach, human experts use an image annotation tool to split each image $I_i$ into $n \times n$ rectangular grids. Then, each grid, $s_{xy}$ (where $1 \leq x,y \leq n$) having corrosion in industrial structures is annotated, producing the grid-based annotated image,  ${I_i}^c$. All the grids in ${I_i}^c$, which are not annotated with corrosion, are considered as non-corrosion grids. The image set produced from original image set $\mathcal{I}$ in this layer  is denoted as $\mathcal{I}^c$  = $\{{I^c}_1, {I^c}_2,...{I^c}_m\}$.

\subsubsection{Object annotation}
Our goal here is to indicate which are parts of industrial component objects given an image, separating them from unimportant noises (e.g. background images - trees) in a given image. This process is known as background separation. By doing so, we desire to focus on identifying corrosion only on the objects related to the target industrial objects. The annotated data of this step are used to learn $\lambda_o$ to detect objects associated with industrial structures.

\begin{figure}[!htb]
    \centering
    \includegraphics[width=3in]{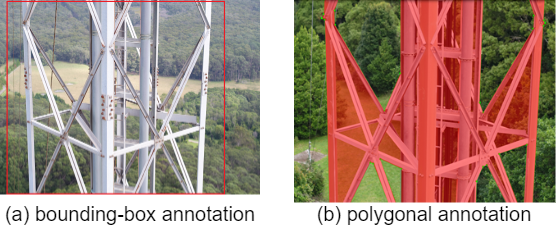}
    \caption{Object annotation methods}
    \label{fig:annotation_process}
\end{figure}

There are two popular annotation approaches used for object separation. The bounding box \citep{russell2008labelme} is the most commonly used approach for such purpose. This basically highlights an object in an image with a rectangular box to make it recognisable for machines. 
The example of this annotation is shown in Figure \ref{fig:annotation_process}(a).
Another approach is polygonal segmentation \citep{russell2008labelme} which is used to annotate objects with irregular shapes, that is, polygons. Unlike bounding boxes,  this approach can exclude unnecessary objects around the target structural objects. Polygons are more precise when it comes to localisation. The example is presented in Figure \ref{fig:annotation_process}(b). 

Our targeted industrial structures have complex and irregular structure and therefore we use the polygonal annotation approach. Given an image, a human expert creates polygonal annotations around each of the target structure objects. Let the annotated image corresponding to a given image $I_i$ be ${I_i}^o$. Let ${\mathcal{M}}_i$ be the polygonal mask (see Figure \ref{fig:annotation_process}(b)) and  ${\mathcal{BBOX}}_i$ be the bounding box information (see Figure \ref{fig:annotation_process}(a)) for ${I_i}^o$. The image set produced after this step from original image set $\mathcal{I}$ is $\mathcal{I}^o$  = $\{{I^o}_1, {I^o}_2,...{I^o}_m\}$.      

\subsection{Processing of annotated images}
Given the image set $\mathcal{I}$, we have now obtained the corresponding grid-based annotated and object annotated image sets in the data preparation layer. We denoted these sets as $\mathcal{I}^c$ and $\mathcal{I}^o$, respectively. These image sets are used to build our CNN models: $\lambda_c$, $\lambda_o$ and final ensemble model $\lambda$ in \model. 

\subsubsection{{Corrosion Image Segmentation and Separation (CISS)} algorithm}


Out of $m$ images in $\mathcal{I}$, $k$ images are randomly sampled to be in a training set, while the rest $m-k$ forms the rest ($k$ is a whole number). The same $k$ images are picked from ${\mathcal{I}}^o$ and ${\mathcal{I}}^c$ where, image ${I_i}^o$ is the object annotated version and image ${I_i}^c$ is grid annotated version for image $I_i$. To develop $\lambda_o$ we use $k$ images in $\mathcal{I}^o$ denoted by, $\mathcal{T}_{\lambda_o}$. Similarly, for  $\lambda_c$ $k$ images used from  $\mathcal{I}^c$ for training is denoted by $\mathcal{T}_{\lambda_c}$.

Given an image $I_i$, its corresponding grid-based annotation image ${I_i}^c$ contains $n \times n$ rectangular grids. Each grid (or segment), $s_{xy}$ is a cropped image with the dimension $w \times h$ (where $w=W/n$ and $h=H/n$) of $I_i$ which have $W \times H$ dimension. The annotated information in ${I_i}^c$ produces a binary matrix $B_i=[ {b}_{xy}]$ of dimension $n \times n$ where $1 \leq x,y \leq n$. The binary decision for $b_{xy}$ is obtained as, 

\begin{equation}\label{eq01}
  \begin{array}{c}
  b_{xy}
  \end{array}
  =
  \left\{
  \begin{matrix}
   1 &  & s_{xy} = \mathcal{L}_c \\
   0 &  & s_{xy} \neq \mathcal{L}_c
  \end{matrix}
  \right.
\end{equation}

That is, $b_{xy}=1$ if the grid of the $x$-th row and $y$-th column is annotated as corrosion, and $b_{xy}=0$, otherwise. Here $\mathcal{L}_c$ represents annotation for corrosion. By applying Equation~\ref{eq01} to all grids in the image ${I_i}^c$, we create the binary matrix $B_i=[ {b}_{xy} ]$. Therefore, for $m$ images in $\mathcal{I}$ (where $m=|\mathcal{I}|$, we obtain the list of $m$ binary matrices, $\mathcal{B}=\{B_1,B_2,..., B_m\}$.

We developed \textit{Corrosion Image Segmentation and Separation (CISS)} algorithm for generating training data, $\mathcal{T}_{\lambda_c}$ for corrosion detection model, $\lambda_c$. The steps of image segmentation and training data separation process is described in Algorithm \ref{alg:image_selection}.

\begin{algorithm}[!htb]
\begin{flushleft}
	\textbf{Input:} {${\mathcal{I}}^c=\{{I_1}^c,{I_2}^c,...{I_i}^c,...,{I_k}^c\}$; $\mathcal{B}=\{b_1,b_2,...b_i,...,b_k\}$}\\
	\textbf{Output:} {Training set for $\lambda_c$, $\mathcal{T}_{\lambda_c}$}
	\BlankLine
	$\mathcal{S}_c \leftarrow \phi$ ;	$\mathcal{S}_{nc} \leftarrow \phi$ \\
	\For{$i$ = 1 to $k$}{
	    $S_i=\{s_{xy}:1 \leq x,y \leq n\} \leftarrow segment({I_i}^c)$\\
        \For {each $s_{xy}$ $\epsilon$ $S_i$}{
            retrieve $b_{xy}$ from $B_i$\\
            \If {$b_{xy}$ = 1}{
                $\mathcal{S}_c \leftarrow \mathcal{S}_c \cup \{s_{xy}\}$
            }
            \Else{
                $\mathcal{S}_{nc} \leftarrow \mathcal{S}_{nc} \cup \{s_{xy}\}$ 
            }
        }
    }
    {
     $N_c  \leftarrow |\mathcal{S}_{c}|$\\
     shuffle all segments $s_{nc}$ $\epsilon$ $\mathcal{S}_{nc}$\\
     ${\hat{\mathcal{S}}}_{nc} \leftarrow $ random 2 $\times N_c$ segments from $\mathcal{S}_{nc}$\\
     $\mathcal{S} \leftarrow \mathcal{S}_c \cup  {\hat{S}}_{nc}$\\
     $\mathcal{T}_{\lambda_c} \leftarrow$ shuffle($\mathcal{S}$)\\
     \textbf{return} ~{$\mathcal{T}_{\lambda_c}$}
    }
    \caption{CISS Algorithm} \label{alg:image_selection}
    \end{flushleft}
\end{algorithm}

In Algorithm \ref{alg:image_selection}, our defined function, $segment(I_i)$ generates $n \times n$ segmented images from $I_i$.  That is,

\begin{equation}\label{eq02}
    I_i \rightarrow S_i=\{s_{xy}:1 \leq x,y \leq n\} 
\end{equation}
    
In total, the number of segmented images produced from $m$ images is $m$$\times$$n$$\times$$n$. 

The segments of all $k$ images are combined and then randomly shuffled. Since the corrosion mainly resides in the industrial structure within an image, the number of identified segments having $\mathcal{L}_c$, $N_c$ is much smaller than the number of non-corrosion segments, $N_{nc}$ (e.g. only 16\% in overall distribution for a real-world use case). To avoid this imbalance issue,  all corrosion segments ($\mathcal{S}_c$) are selected for model development. Then a random samples of non-corrosion segments ($\mathcal{S}_{nc}$) are chosen such that the selected number of non-corrosion segments $N_{nc}$ = 2 $\times$ ${N}_c$. This produces the final non-corrosion image segment set $\hat{S}_{nc}$. In the end, $\mathcal{S}_c$ and $\hat{S}_{nc}$ are combined and shuffled which produces the final image set $\mathcal{T}_{\lambda_c}$ which is used as the input image set for $\lambda_c$


\subsection{\model CNN models development}

The \model framework incorporates the following models that have been developed for detecting corrosion in images of industrial structures. In the rest of this section, we provide detailed descriptions of the  model development process.  

\begin{itemize}
     \item $\lambda_c$: Given the training set, $\mathcal{T}_{\lambda_c}$ containing random $ 3 \times N_c (N_c + 2 \times N_c) $ image segments from  $\mathcal{I}^c$.  Each segment $s_{xy}$ belongs to $\mathcal{T}_{\lambda_c}$ as described in Algorithm \ref{alg:image_selection}. The objective of $\lambda_c$ is to automatically identify segments $s_{xy}$ having $\mathcal{L}_c$ in each of $m-k$ images in $\mathcal{I}$ and compare with the ground truth $\mathcal{L}_c$ in same $m-k$ images in $\mathcal{I}^c$.
     
    \item $\lambda_o$: Given the training set, $\mathcal{T}_{\lambda_o}$ containing $k$ images from $\mathcal{I}^o$. Each ${I_i}^o \epsilon {I}^o$ is annotated with target object $T$. The objective of $\lambda_o$ is to automatically estimate $\mathcal{\hat{M}}_{o}$ in $m-k$ images in $\mathcal{I}$ and compare with the ground truth $\mathcal{M}_{o}$ in same $m-k$ images in $\mathcal{I}^o$.
    
    \item $\lambda$: Given the outcome of $\lambda_c$ and $\lambda_o$. The objective of ensemble model, $\lambda$ is to estimate corrosion for each individual segments $s_{xy}$ in a test image $I_i$ such that $s_{xy}$ is detected as a corrosion segment by $\lambda_c$ and the presence of targeted object is detected in $s_{xy}$  by $\lambda_o$.
\end{itemize}


\subsubsection {\textit{CNN model for corrosion detection in image segment-level, $\lambda_c$}}
\label{sec:corr_model}

Our proposed CISS algorithm generates multiple rectangular segments ($s_{xy}$) of the complete image. Each segment represents a smaller size image. Therefore, for detecting corrosion in such small image, the CNN mainly needs to learn the features for the corrosion color. A simple CNN can serve such purpose. 

In general, a CNN consists of multiple convolution (CO) and pooling layers followed by the fully connected (FC) and classification layers. CO are used to extract features from the training images. CO consists of kernels (a set of small receptive fields). The weight values for the kernels are typically initialised with random values and updated during training. Pooling layers are used to decrease the data to decrease the computational costs during the training phase. There are two well-known methods in pooling: (1) max pooling (MP) and (2) mean pooling (MeP). 

The goal of designing a simple CNN is to determine whether such CNN model can be used to predict corrosion in a good performance. Thus, to construct $\lambda_c$, we propose a shallow version of VGG16 \citep{simonyan2014very} with 5 layers - 3 convolution (CO) layers and 2 fully-connected (FC) layers (see Table \ref{tab:corrcnn_net}. Our model is inspired by the approach proposed in \citep{atha2018evaluation} where they proposed a 7-layer CNN called Corrosion7 (see Table \ref{tab:corrosion7_net}). 

\begin{table}[!htb]
	\centering
	\caption{$\lambda_c$ architecture - CO:convolution, MP: Max Pooling, KS: Kernel Shape, NK: Number of Kernels, NV: Number of Variables. Each number indicate a layer number.}
 
  \begin{tabular}{|l|l|l|l|l|l|}  
  \hline
  Layer & Input  & MP & KS  & NK & NV\\
  \hline
  CO1 & 224 X 224 X3 & 2 X 2 & 3 X 3 & 32 & 288\\
  \hline
  CO2 & 224 X 224 X 32 & 2 X 2 & 3 X 3 & 64 & 576\\
  \hline
  CO3 & 64 X 64 X 64 & 2 X 2 & 3 X 3 & 128 & 1152\\
  \hline
  FC1 & 2 X 2 X 128 & - & 512 & 128 & 65536\\
  \hline
  FC2 & 128 & - & 128 & 2 & 128\\
  \hline
  \end{tabular}
 \label{tab:corrcnn_net}
\vspace{5pt}
	\centering
	\caption{Corrosion7 architecture}
  \begin{tabular}{|l|l|l|l|l|l|}  
  \hline
  Layer & Input  & MP & KS  & NK & NV\\
  \hline
  CO1 & 128 X 128 X 3 & 64 X 64 & 3 X 3 & 58 & 8526\\
  \hline
  CO2 & 33 X 33 X 58 & 17 X 17 & 3 X 3 & 128 & 185600\\
  \hline
  CO3 & 9 X 9 X 128 & 9 X 9 & 3 X 3 & 192 & 221184\\
  \hline
  CO4 & 9 X 9 X 192 & - & 3 X 3 & 192 & 331776\\
  \hline
  CO5 & 9 X 9 X 192 & - & 3 X 3 & 128 & 221184\\
  \hline
  FC1 & 4 X 4 X 128 & - & 512 & 1024 & 2097152\\
  \hline
  FC2 & 1024 & - & 128 & 2 & 2048\\
  \hline
  \end{tabular}
\label{tab:corrosion7_net}
\end{table}

After constructing $\lambda_c$ the following steps are performed,

\begin{enumerate}
    \item $\lambda_c$ is trained with the training set, $\mathcal{T}_{\lambda_c}$  generated in CISS Algorithm (Algorithm \ref{alg:image_selection}).
    \item The trained $\lambda_c$ is tested with $m-k$ images in $\mathcal{I}^c$. 
    \item For a given test image, $I_i$  $\epsilon$ $\mathcal{I}$, $\lambda_c$ generates decision for $\mathcal{L}_c$ using confidence value $conf(s_{xy})$ for each $n \times n$ rectangular segments, $s_{xy}$ in $I_i$ where $ 0 \leq conf(s_{xy}) \leq 1$.
    \item  Let $\mathcal{CS}_i=[ conf(s_{xy})]$ be the confidence matrix for $I_i$, for $m-k$ images $\lambda_c$ generates, $\mathcal{CS}=\{\mathcal{CS}_1,\mathcal{CS}_2,...\mathcal{CS}_i,...\mathcal{CS}_{m-k}\}$.
    \item  The decisions for $\mathcal{L}_c$ for all segments $s_{xy}$ in $I_i$  produce binary matrix $\hat{B}_i=[ {\hat{b}}_{xy} ]$ where each ${\hat{b}}_{xy}$ is computed according to Equation \ref{eq01}. We call this segment-level prediction (SLP). $\mathcal{\hat{B}}=\{{\hat{B}}_i, {\hat{B}}_2,...{\hat{B}}_i,...{\hat{B}}_{m-k}\}$ .
    \item The overall confidence for image $I_i$ is computed as,

\begin{equation}
    conf_c(I_i) =\frac{\sum_{x=1}^{n}\sum_{y=1}^{n}{\hat{b}}_{xy}}{n \times n}
\end{equation}

\end{enumerate}

These confidence values, \\$\mathcal{CC}=\{conf_c(I_1), conf_c(I_2),...conf_c(I_i),...conf_c(I_{m-k})\}$ are used to make decision for image-level (IL) prediction for corrosion.

\subsubsection {\textit{CNN model for Industrial object identification, $\lambda_o$}}

We have used Mask R-CNN \citep{he2017mask} for developing the model for object identification, $\lambda_o$. This is one of the best performing model in object detection and instance segmentation \citep{he2017mask}. For a given image, $I_i$, this network can separate different objects. In our industrial object identification problem we have only one target object. Mask R-CNN provides output for the object bounding boxes, classes and masks. There are two stages of Mask R-CNN (region proposal network and neural network) and both of them are connected to a backbone network. 

The backbone network for Mask R-CNN is a deep neural network which is responsible for extracting features from raw image. The deeper network may result higher accuracy, however, can highly impact on the duration of  model training and classification. By passing through backbone network, images are converted into feature maps. A top-down pyramid structure, Feature Pyramid Network (FPN) \citep{lin2017feature} is  used to extract features. The extracted features from top layers are transferred to lower layers. Due to this structure each layer in pyramid has access to the higher and lower layers. In this context, we have used RestNet-101 FPN backbone as feature extractor for our Mask R-CNN to increase its speed and performance. 

The first stage of Mask R-CNN is a light weight neural network called (RPN) that generates regions of interest (RoIs) from feature maps provided by backbone network \citep{he2017mask}. The second stage is another neural network takes proposed RoIs by the first stage and assign them to several specific areas of a feature map level, scans these areas, and generates objects, bounding boxes and masks. There is a branch for generating masks for each objects in pixel level. It also provides the confidence value of detected bounding box similar to Faster R-CNN.

After constructing the Mask R-CNN model with backbone network ResNet-101, $\lambda_o$, the following steps are performed.

\begin{enumerate}
    \item $\lambda_o$ is trained with training set, $\mathcal{T}_{\lambda_o}$ containing $k$ images from $\mathcal{I}^o$.
    \item The trained $\lambda_o$ is then tested with remaining $m-k$ images in $\mathcal{I}^o$.
    \item For a given test image, $I_i$  $\epsilon$ $\mathcal{I}$, $\lambda_o$ generates output of the estimated mask of target object, $\mathcal{\hat{M}}_{o_i}$ with a confidence value $conf_o(I_i)$. $conf_o(I_i)$ is used to make decision for object detection for $I_i$ and $ 0 \leq conf_o(I_i) \leq 1$.
    \item The test result for all $m-k$ images in $\mathcal{I}$ finally generates 
    the list of images containing $\mathcal{\hat{M}}_{o_i}$ estimated by $\lambda_o$,  ${\mathcal{\hat{I}}^o}_{m-k}$ and the list of their confidence values , $\mathcal{CO} = \left [ conf_o(I_i)  \right ]_{i=1}^{m-k}$
\end{enumerate}

\subsubsection{Ensemble model for region-based corrosion detection, $\lambda$}
\begin{figure}[!htb]
    \centering
    \includegraphics[width=5in]{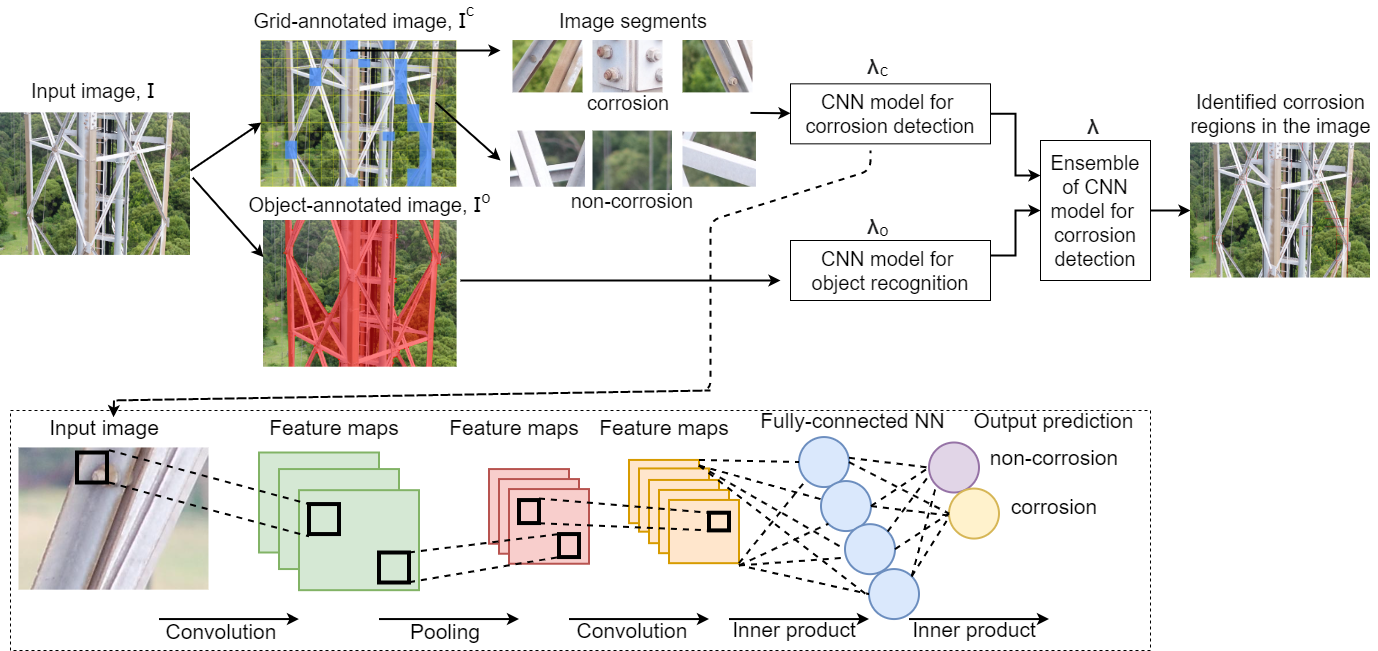}
    \caption{An ensemble DL model with an overview CNN structure for corrosion detection}
    \label{fig:cnn}
\end{figure}

The Ensemble model, $\lambda$ is a machine learning model that utilises the outcome from $\lambda_c$ and $\lambda_o$ to make the final decision for corrosion in an image segment (i.e. region of an image). Figure \ref{fig:cnn} shows the process of development of the ensemble model $\lambda$ from the outcomes of $\lambda_c$ and $\lambda_o$. An example of a CNN structure for corrosion detection in an image segment is also presented in the bottom part of the figure.

Algorithm \ref{alg:erc} outlines the steps involved in our proposed Ensemble Region-based Corrosion (ERC) detection.

\begin{algorithm}[!htb]
	\begin{flushleft}
	\textbf{Input:} ${\mathcal{\hat{I}}^o}_{m-k}$; $\mathcal{CO}$;   $\mathcal{B}$;$\mathcal{\hat{B}}$; $\mathcal{CS}$\\
	\textbf{Output:} {Training set for $\lambda$: $FC$ and $FB$}\\
	
	$FB \leftarrow \phi$; $FC \leftarrow \phi$\\
	\For {each ${\hat{I_i}}^o$ $\epsilon$ ${\mathcal{\hat{I}}^o}_{m-k}$;
	$B_i$ $\epsilon$ $\mathcal{B}_{m-k}$; ${\hat{B}}_i$ $\epsilon$ $\mathcal{\hat{B}}$; $conf_o(I_i)$ $\epsilon$ $\mathcal{CO}$; ${\mathcal{CS}}_i$  $\epsilon$  $\mathcal{CS}$ }{
	    ${\Bar{B}}_i \leftarrow flatten(B_i)$; 
	    ${\Bar{\hat{B}}}_i \leftarrow flatten(\hat{B}_i)$; 
	    ${\Bar{CS}}_i \leftarrow flatten({\mathcal{CS}}_i)$\\
	    $ \mathcal{\hat{M}}_{o_i} \leftarrow mask({\hat{I_i}}^o) $;
	    $A_i = area(\mathcal{\hat{M}}_{o_i})$\\
	    $S_i=\{s_{xy}:1 \leq x,y \leq n\} \leftarrow segment({\hat{I_i}}^o)$\\
        \For {each $s_{xy}$ $\epsilon$ $S_i$}{
            $intersectVal \leftarrow area(s_{xy}) \cap A_i$\\ 
            \If {$intersectVal \geq$ 10\%}{
                $tB_{xy} \leftarrow$ 1\\
                $tI_{xy} \leftarrow intersectVal \times conf_o(I_i)$
            }
            \Else{
                $tB_{xy} \leftarrow$ 0; $tI_{xy} \leftarrow$ 0
            }
        }
        $\Bar{tB}_i \leftarrow flatten(tB_{xy})$; $\Bar{tI}_i \leftarrow flatten(tI_{xy})$\\
        ${fc}_i \leftarrow \{{CS}_i,{tI}_i,B_i\}$;
        ${fb}_i \leftarrow \{{\Bar{\hat{B}}}_i,{tB}_i,B_i\}$\\
        $FC \leftarrow FC \cup {fc}_i$; 
        $FB \leftarrow FB \cup {fb}_i$ 
    }
    {
     \textbf{return} ~{$FC$, $FB$}
    }
    \caption{ERC Algorithm} \label{alg:erc}
    \end{flushleft}
\end{algorithm}

The steps of Algorithm \ref{alg:erc} are discussed as follows

\begin{itemize}
    \item The algorithm generates 2 different feature set, $FC$ and $FB$, using confidence values and binary decision values of each predicted image segment $s_{xy}$ by $\lambda_c$ respectively.
    \item  The $flatten(X)$ function converts $n \times n$ matrix $X$ to $n^2$ elements single dimensional array.
    \item  To convert the mask outcome of $\lambda_o$, $\mathcal{M}_o$ to a binary matrix similar to the outcome of $\lambda_c$. We first convert the masked outcome of image, $I_i$ into same $n \times n$ segments. Then we compute whether a segment belongs to the detected object using the area of intersection of the segment and object mask.
    \item  We consider a segment is part of the object, if object mask overlap with more than 10\% area of the segment. If the condition is true we assign value 1 and 0, otherwise. Here, we ignore a small amount of of overlap (10\%) based on the observation that if that part of this object contains corrosion that should be detected in remaining 90\% overlapped with other segments.
    \item Finally, we combine this binary outcome of $\lambda_o$ with $\lambda_c$ and true label which produce the feature set, $FB$. 
    \item Another feature set $FC$ is generated based on confidence value of object mask. If the segment completely overlap with object mask then the confidence value of the segment is same like object mask. Otherwise, it is the fraction of overlapped area. In case of less than 10\% overlap the confidence value is 0. This outcome is combined with confidence values of each segment for all test images along ground truth label for corrosion in $FC$.
\end{itemize}

The generated feature set $FB$ is then fed into a machine learning classifier such as: support vector machine (SVM), Multilayer Perceptron (MLP) and XGboost using binary outcome of $\lambda_c$ and $\lambda_o$ as features and true values as class. Similarly, $FC$ is also fed similar machine learning classifiers using confidence values of 2 models as features and ground truth corrosion value as class.

\subsection{Decision for corrosion}
As stated before, using trained models several prediction decision for corrosion are made in \textit{decision layer}. They are stated as follow.

\begin{itemize}
    \item \textbf{Segment-level prediction (SLP)}: Using $\lambda_c$ we predict corrosion or non-corrosion of the segmented images in the test set. This prediction is done using the confidence value $conf(s_{xy})$ for segment $s_{xy}$ estimated by $\lambda_c$. If $conf(s_{xy}) \geq \tau_s$ then, segment has corrosion, otherwise, non-corrosion. Here, $\tau_s$ is the minimum confidence value to satisfy the condition for a segment has corrosion and $0 \leq \tau_s, conf(s_{xy}) \leq 1$.
    \item \textbf{Image-level prediction (ILP)}: This means a corrosion prediction for each test image by {\it aggregating} its segment-level predictions from $\lambda_c$. An image. $I_i$ should have $conf_c(I_i) \geq \tau_I$ to satisfy the condition or corrosion. Where $\tau_I$ is the fraction of minimum number of segments should have corrosion among all segments in $I_i$. $\tau_i$ has been determined as the the mean value of proportion of corroded segments in all segments in the train set and  $0 \leq \tau_I, conf(s_{xy}) \leq 1$.
    \item \textbf{Industrial object  prediction (IOP)}: Using $\lambda_o$ we predict the occurrence  and position of target industrial object in test images. This prediction is done using the confidence value $conf_o(I_i)$ for image $I_i$ estimated by $\lambda_o$. If $conf_o(I_i) \geq \tau_o$ then, image contains the target object, otherwise, false. Here, $\tau_o$ is the minimum confidence value to satisfy the condition for occurrence of target object in the image and $0 \leq \tau_o, conf_o(I_i) \leq 1$.
\end{itemize}

\section{Evaluation}
\label{sec:evaluation}

We evaluate \model on real-world images captured by drones. These images are unaltered and used directly in our experimentation. None of existing work in literature used drone images for corrosion detection. Most of these images are collected from web resources \citep{bonnin2014corrosion,hoskere2018vision} or altered/cropped version of images captured by digital camera \citep{atha2018evaluation, petricca2016corrosion, fernandez2013automated}.

The objective of our evaluation is to measure the performance of the ensemble model $\modelname$ in the \model framework, in comparison with some state-of-the-art CNN models used for corrosion identification.


\subsection{Evaluation Domain}
We focus on detecting corrosion from telecommunication towers. Thus, we have collected the images of telecommunication towers captured using camera installed in drones. 
The drones are controlled by operators and they randomly capture different views of telecommunication tower. This differs from many prior works that have conducted evaluations on images collected in controlled settings (i.e. only for the purpose of experimental evaluation) \citep{atha2018evaluation,cha2018autonomous}. 
     
\begin{figure}[!htb]
    \centering
    \includegraphics[width=5in]{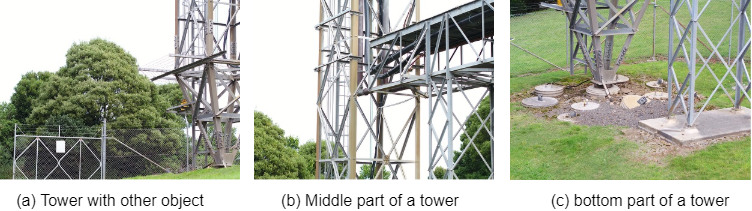}
    \caption{Variations in tower structure in images captured by drones}
    \label{fig:complex_structure}
\end{figure}

We have used a total of 573 high-resolution images that contain different views under various lighting conditions of telecommunication towers (see an example in Fig.~\ref{fig:complex_structure}). Each image contains a single tower object and has a $5280 \times 3952$ resolution. 
A telecommunication tower has a complex structure. When random images are captured about it by drones, a high variations is observed in the images (e.g. middle part in Figure \ref{fig:complex_structure}(b) and bottom part in  Figure \ref{fig:complex_structure}(c)). Also, as seen in \ref{fig:complex_structure}(a), some part of tower images are overlapped with other objects (e.g. gate) Therefore,  it is hard to distinguish which part of the telecommunication tower needs to be inspected for corrosion detection. However, \model can detect such a part correctly and identify the state of corrosion on it.

\subsection{Dataset preparation}
All the captured images are annotated by experts. Grid-based annotations are performed using the RatSnake \citep{iakovidis2014ratsnake} software. RatSnake is an annotation tool that is capable of fast annotation of images with polygons, grids or both. Image annotations produced by RatSnake can be exported to various formats. In the grid-based annotation approach, we used $n$=16, that is, each image is segmented and cropped equally to $16 \times 16$ segments. Each segment ends up with having a $330 \times 247$ resolution (see examples in Fig.~\ref{fig:sample_corrosion}). Thus, we have a total of 146,688 segmented image samples (573$\times$16$\times$16). 

We have also used object annotation for background separation. Object annotation is done using Labelme \citep{russell2008labelme}. LabelMe is a popular annotation tool, where a user can draw both bounding boxes and set of polygon points for segmentation maps. Both online and desktop version of this tool is available. The points of annotated polygon can be saved in JSON or xml format which are used as input for CNN model.

For background separation, authors in \citep{atha2018evaluation} first manually cropped the portion of the image having industrial structure. Then, it used that cropped image for segmentation into three different regions, 128$\times$128, 64$\times$64, 32$\times$32. However, in our approach, we have used a $16 \times 16$ segmentation on an original image since telecommunication tower views have a complex structure (as in Figure \ref{fig:complex_structure}). Moreover, the structure of tower is not simple in contrast to a metallic surface as the presence of background can be present within image area covered by the tower. Therefore, it is difficult and time consuming to separately crop only part having corrosion that is proposed in \citep{atha2018evaluation}.  

Our choice of the 16$\times$16 segment size is intuitive as popular CNN models have often used a $224 \times 224$ resolution as input image shape \citep{gulli2017deep}. By being scaled at a  $224 \times 224$ resolution from the $330 \times 247$ resolution image segment, we are able to use the image samples to train $\lambda_c$ nearly in their original resolutions and avoid high distortion due to down-scaling (i.e. from $330 \times 247$ scaled to $224 \times 224$). Also, the number $16$ is chosen to be the maximum granularity level value where we can get an integer number in $h$ (height of segmented image in terms of pixel) which is above $224$ and so we can segment according to pixel position in image. RGB (3-channels) colour space is used for forming the input shape of CNN. 


\subsection{Separation of training and test set}
To build any supervised machine learning model, an essential step is to split sample data into train and test sets. Train set is used to build CNN models whereas the test set is utilised for validating the models. 

\begin{figure}
\centering
\begin{subfigure}{.5\textwidth}
    \centering
	\includegraphics[width=2.3in]{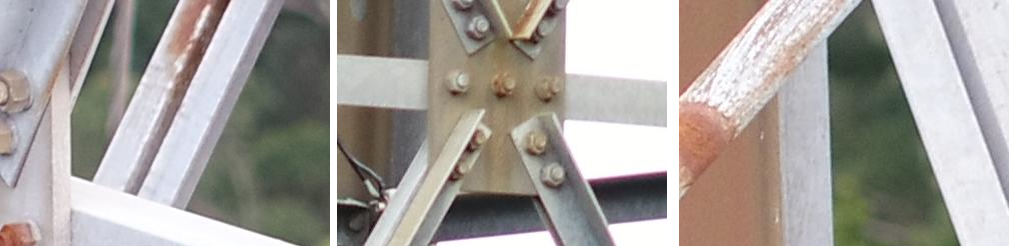}
	\caption{Corrosion segments}
	\label{fig:sample_corrosion}
\end{subfigure}%
\hfill
\begin{subfigure}{.5\textwidth}
 \centering
    \includegraphics[width=2.3in]{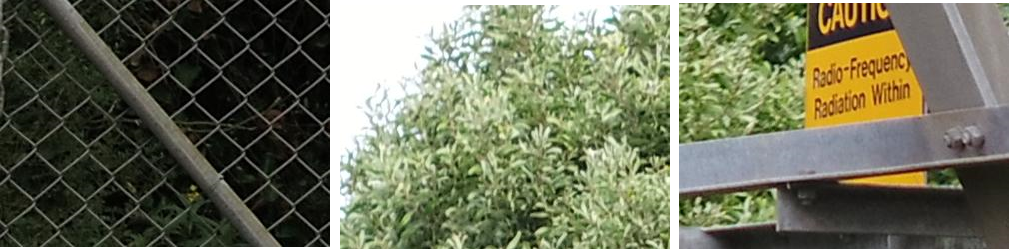}
	\caption{Non-corrosion segments}
	 \label{fig:sample_noncorrosion}
\end{subfigure}
\caption{Example of Corrosion and non-corrosion segments}
\end{figure}

Given the $m$=573 images, we used $k$=379 images (approx. $\frac{2}{3}$ of $573$) for training the CNN models for $\lambda_c$. A total of the 15,863 segmented images are marked as corrosion by the experts out of the 97,024 (379$\times$16$\times$16) segments. Three such sample images are shown in Figure \ref{fig:sample_corrosion}. To avoid a class (i.e. corrosion and non-corrosion) imbalance problem, a total of the 31,726 samples out of 81,161 (97,024$-$15,863) were randomly chosen as non-corrosion samples. Thus, we used a total of 47,589 (15,863$+$2$\times$15,863) image segments for training as described in Algorithm \ref{alg:image_selection}. We used $80\%/20\%$ as train/validation set split. Overall, we have a total of 38,701 images for training and 8,888 images for validation in each iteration of CNN model.

During the human annotation phase, as only corrosion segments within the tower structure are annotated by the experts, everything else outside those segments (including background such as trees, grounds, sky and other objects of images, etc) are considered non-corrosion. Some non-corrosion sample segments are presented in Figure \ref{fig:sample_noncorrosion}. 

To evaluate the performance of $\modelname_c$, we used 194 ($m-k=573-379=194$) images not used for training. That is, we measured the performance of models developed for detecting corrosion in image segment-level using 49,664 image segments (194$\times$16$\times$16).

To train $\lambda_o$, annotated polygonal mask of $m$=379 images are used for training Mask R-CNN model. The model is evaluated using the remaining 194 images. During the training, we used $70\%/30\%$ split that is, 265 image for training and 114 images for validation. Here we used ResNet101 as backbone network (ResNet101 is 101 layers deep). We prefer this split over $80\%/20\%$ which we used for $\lambda_c$ so we can get higher number of images in validation.

\subsection {Evaluation Metrics}
We evaluate the performance of individual CNNs and the ensemble model, $\modelname$, using different performance measures over test set. We use accuracy ($Acc.$), precision ($P$), recall ($R$) and F1-score ($F1$), obtained from true positive ($TP$), false positive ($FP$),  true negative ($TN$) and false negative ($FN$) values from the classification result (i.e. confusion matrix) over test set:  
\begin{align*}
\end{align*}

To evaluate $\lambda_o$ we also measure Intersection over Union (IoU) \citep{rezatofighi2019generalized} for $I_i$ as in equation \ref{eq:iou}.
\begin{equation}
\label{eq:iou}
    IOU(I_i) = \frac{Area(\mathcal{\hat{M}}_{o_i} \cap \mathcal{M}_{o_i} )}{Area(\mathcal{\hat{M}}_{o_i} \cup \mathcal{M}_{o_i})} 
\end{equation}
This computes the amount of overlap between the detected and the ground truth polygonal masks. Similar measure is defined for bounding box annotation also (Equation \ref{eq:iou_bb}).

\begin{equation}
\label{eq:iou_bb}
    IOU(I_i) = \frac{Area(\mathcal{\hat{BBOX}}_{o_i} \cap \mathcal{BBOX}_{o_i} )}{Area(\mathcal{\hat{BBOX}}_{o_i} \cup \mathcal{BBOX}_{o_i})} 
\end{equation}

\begin{figure}[!htb]
    \centering
    \includegraphics[width=4in]{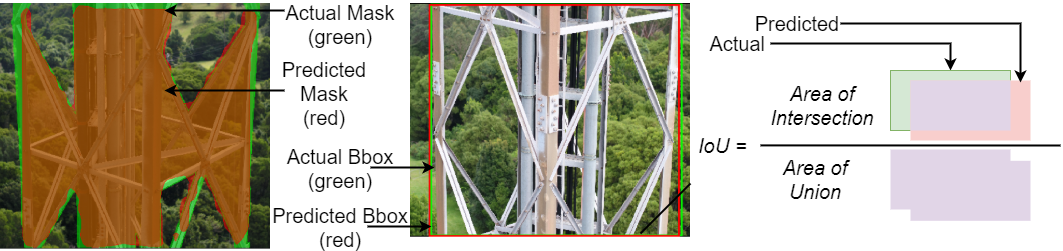}
    \caption{The process of computing \textit{IoU} using ground truth and predicted value}
    \label{fig:iou}
\end{figure}

Figure \ref{fig:iou} shows the demonstration for computing \textit{IoU} for mask and Bbox. 

To measure the performance of object position detection in $I_i$, if $IOU(I_i) \geq IOU_{th}$ then it is considered that the target object mask $\mathcal{\hat{M}}_{o_i}$ estimated the position in the object $T$ in $I_i$ accurately (true positive ($TP$) case).  Otherwise, this is considered as false positive ($FP$). Here $0 \leq IOU_{th} \leq 1$ and generally, $TP$ is considered when $ IOU_{th} > 0.5$. Thus, using $TP$, $FP$ it is possible to compute precision using the outcome from all test images. 

Object detectors normally use Average precision(AP) as metric to measure the performance of the detection. AP is the average over multiple $IOU_{th}$. For example, AP for $IOU_{th}$ values from 0.5 to 0.75 with a step size of 0.05. We also used AP for for evaluating $\lambda_o$ using $IOU$ values for object mask and bounding boxes. 

We use human annotations as gold standards for computing accuracy across all models. 

\subsection {Compared State-of-the-art CNN Models}
Our ensemble model $\modelname$ is also compared with some state-of-the-art deep learning models described in Section \ref{sec:literature}. In particular, we compare $\lambda_c$ with two models (i.e. Corrosion5 and Corrosion7) specifically designed for corrosion detection in \citep{atha2018evaluation} using cropped image. Further, we compare $\lambda_c$ with  four pre-trained models through transfer learning, InceptionV3 \citep{sam2019offline}, MobileNet \citep{howard2017mobilenets}, Rsetnet50 \citep{he2016deep},  and Vgg16 \citep{simonyan2014very}. Corrosion5 is a simpler version of Corrosion7 described in Table \ref{tab:corrosion7_net} with 3 convolution layers and 2 fully connected layers. Corrosion5 and Corrosion7 in \citep{atha2018evaluation}, used different kernel shapes and authors have not clearly explained the reason for choosing such kernel shapes. Our model, $\lambda_c$ as described in Table \ref{tab:corrcnn_net} is a 5 layered CNN with the first FC has 128 neurons followed by the binary classification layer (i.e. corrosion or not) with 2 neurons. To evaluate how corrosion detection performs with a simpler structure, we also develop a simple CNN with only one convolution layer and two fully-connected layers.

\subsection {Evaluation of developed models: $\lambda_c$, $\lambda_o$ and $\lambda$}

The two CNN models $\lambda_c$ and $\lambda_o$ are trained and evaluated in multiple iterations and various configurations on two NVIDIA Tesla P100-PCIE-12GB GPUs on CUDA with four Intel Gold 6140 18-core processors. Finally, $\lambda$ is developed using the outcome of $\lambda_c$ and $\lambda_o$. In this section, the model development process and evaluation outcomes are described.

\subsubsection{Training and Evaluation for Corrosion Detection Model $\lambda_c$}
As stated in section \ref{sec:corr_model}, $\lambda_c$ is developed to predict corrosion or not as a binary classification problem. Thus, the binary cross-entropy is used as a loss function. 

To train $\lambda_c$, different configurations of hyper-parameters are used such as input dimensions for sample images (e.g. $128 \times 128$, $200 \times 200$, $220 \times 220$, $224 \times 224$), batch size (e.g. 16, 32, 64, 128) and number of epochs (10, 20, 30, 50, 100). Moreover, rotation, shear range, zoom range, width and height shift and horizontal flips are used during input data generation \citep{gulli2017deep}. 

The model with the best outcome in terms of performance measure (e.g. low training loss) are retained. Finally, all the compared CNN models are trained using same configurations (i.e. batch size=64, number of epoch =30 and input image shape 224$\times$224). For the transfer learner models, the pre-trained weights are fine-tuned to re-train with generated training set. 

\begin{figure}[!htb]
    \centering
    \includegraphics[width=3.5in]{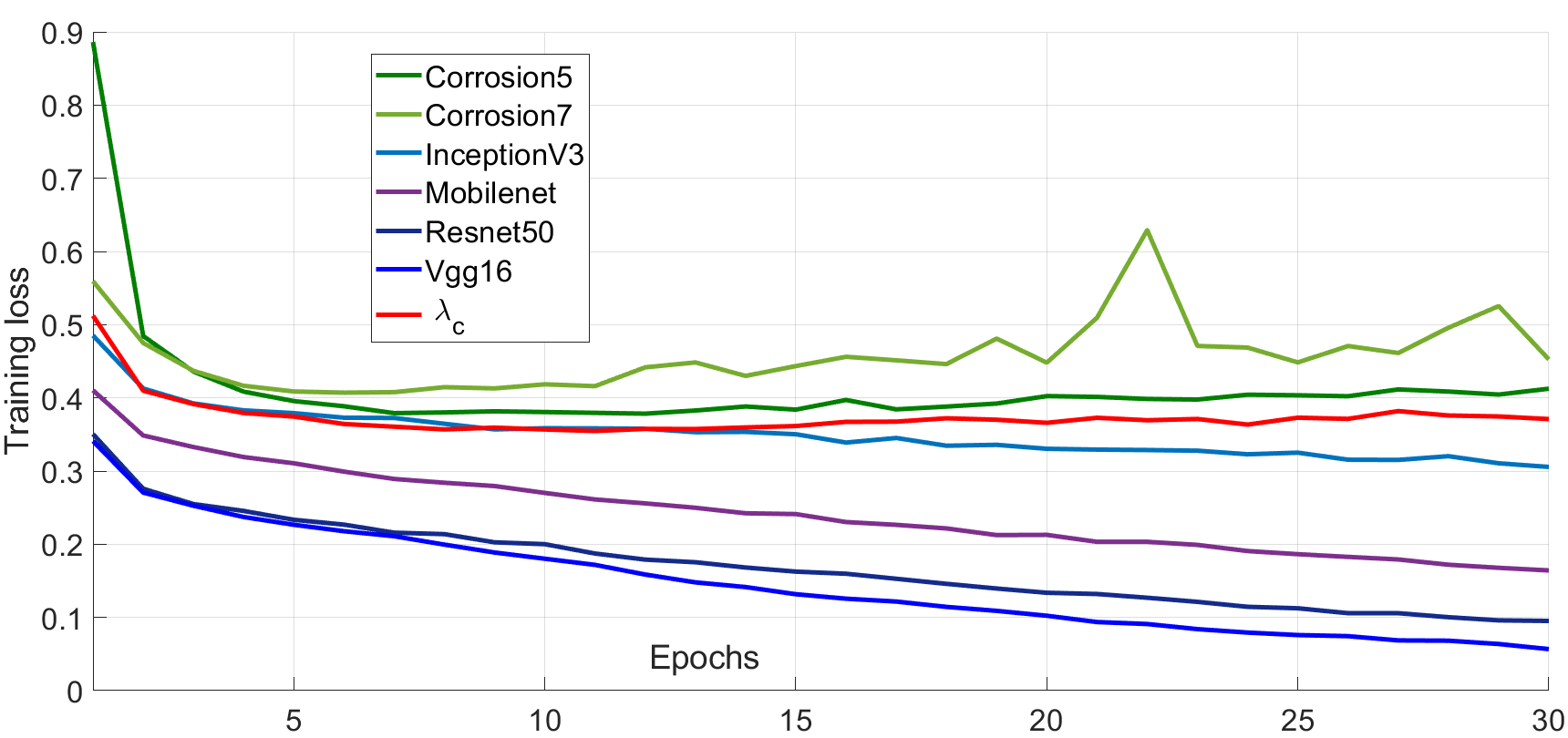}
    \caption{Training loss for the seven CNN models compared}
    \label{fig:training_loss}
\end{figure}

The learning curves based on the the same loss function across eight models during training are presented in Figure \ref{fig:training_loss}. As observed, all the models were converged after around 30 epochs. The training loss was lower in four pre-trained models - InceptionV3, MobileNet, Restnet50 and VGG16. Our model, $\lambda_c$, showed stability than Corrosion5 and Corrosion7. The training loss was higher in Corrosion5 and Corrosion7 \citep{atha2018evaluation} than $\lambda_c$ and four pre-trained models. Moreover, Corrosion5 had a high loss value at the beginning and Corrosion7 was unstable (as shown between epoch 20 and 30). The loss value for $\lambda_c$  after 30 iterations was 0.3702. The pre-trained model converged very quickly and the training losses are lower than $\lambda_c$. The lowest loss after 30 iterations was 0.0567 obtained in Vgg16. However, though the pre-trained models performed well in training data, lower accuracy were observed on test data (discussed below).

In the evaluation, we first measure the performance of $\lambda_c$ without integrating $\lambda_o$. Due to variations in tower images and corrosion from unknown sources the evaluation outcome from $\lambda_c$ can generate many false positives (FPs). The reason of building $\lambda_o$ is to suppress those FPs so that detected corrosion segments remain in tower object. Therefore, each of image is annotated separately for building $\lambda_c$ and $\lambda_o$. We have used $\tau_s = 0.5$ (for $\lambda_c$) as it is a general standard value for CNN in binary classification \citep{cha2018autonomous}. $\tau_s$ is minimum confidence value to satisfy the condition for a segment, $s_{xy}$ has corrosion and $0 \leq \tau_s, conf(s_{xy}) \leq 1$.

\begin{table}[!htb]
  \caption{Performance comparison for segment-level and image-level prediction}
  \label{tab:evaluation}
  \centering
  \begin{tabular}{|p{3cm}|l|l|l|l|l|l|l|l|l|}
    \hline
    \textbf{CNN model} & \multicolumn{2}{c|} {\textbf{Acc.(\%)}} &  \multicolumn{2}{c|}{\textbf{P(\%)}} & \multicolumn{2}{c|}{\textbf{R(\%)}} & \multicolumn{2}{c|}{\textbf{F1(\%)}} \\
    \hline
    & \textbf{SLP} & \textbf{ILP} & \textbf{SLP} & \textbf{ILP} & \textbf{SLP} & \textbf{ILP} & \textbf{SLP} & \textbf{ILP}\\
    \hline
    Simple CNN  & 58.17 & 37.52  & 22.31 & 38.36  & 34.58 & 88.41  & 27 & 54\\
    \hline
    \cellcolor{yellow!25}Our model, $\lambda_c$ & \cellcolor{yellow!25}86.28 & \cellcolor{yellow!25}92.50  & \cellcolor{yellow!25}53.42 & \cellcolor{yellow!25}96.01  & \cellcolor{yellow!25}85.41 & \cellcolor{yellow!25}95.91 & \cellcolor{yellow!25}66 & \cellcolor{yellow!25}98 \\
    \hline
    Corrosion5 \citep{atha2018evaluation}  & 78.24 & 62.48  & 65.67 & 65.92  & 44.7 & 91.7  & 53 & 77\\
    \hline
    Corrosion7 \citep{atha2018evaluation}  & 81.67 & 68.41 & 60.02 & 70.76  & 52.60 & 94.05 & 56 & 80 \\
    \hline
    InceptionV3 & 81.58 & 66.31  & 42.34 & 69.10  & 63.93 & 92.94  & 51 & 79\\
    \hline
    Mobilenet  & 73.86 & 61.76 & 43.22 & 64.30 & 25.18 & 92.47  & 32 & 76 \\
    \hline
    Resenet50  & 75.04 & 63.33  & 44.71 & 65.61  & 31.23 & 93.13  & 37 & 77 \\
    \hline
    Vgg16  & 77.97 & 68.88  & 51.81 & 70.20  & 58.96 & 94.01 & 55 & 80 \\
    \hline
\end{tabular}
\end{table}

The evaluation results in terms of the evaluation metrics (i.e accuracy, precision, recall and F1-score) on segment-level prediction (SLP) and image-level prediction (ILP) are presented in Table \ref{tab:evaluation}. As seen, our model, $\lambda_c$, already outperforms the other seven CNN models in terms of all the performance metrics (highlighted in yellow) except precision in SLP. The simple CNN performs the worst which confirms that a single CNN layer may not be sufficient for corrosion detection. In the pre-trained models, the accuracy of InceptionV3 is turned to be better than the others but Vgg16 is better in terms F1 due to the higher precision value. 

In terms of accuracy, recall and F1-score $\lambda_c$ is also the best. The two models, Corrosion5 and Corrosion7, in \citep{atha2018evaluation} did not perform well on our test data. Four pre-trained models did not perform so well in test data even though they were outstanding in train data. 

The prediction results of image $I_i$, matrix $B_i$  (i.e SLP) in segment-level are aggregated to detect corrosion for ILP. Out of 573 annotated images used in training and testing in total 10\% segments are marked as corrosion by the human experts. Therefore, we used the threshold, $\tau_I$ = 0.1. That is, given an image, if more than 10\% segments are marked as corrosion by a CNN model, then this image is considered to be corrosion. Using this notion, the performance of ILP for 194 images are calculated which is presented in Table \ref{tab:evaluation}. As seen, $\lambda_c$ showed 92.5\% in Accuracy and 98\% in F1-score. This shows the effectiveness of our proposed model over the other state-of-the-art and pre-trained models we compared with. However, due to high number of false positives (FPs) the precision in SL was low. To improve this and eliminate FPs happened outside tower structure, we developed $\lambda_o$ and then combined it in ensemble model $\lambda$.

\subsubsection{Training and Evaluation for Object Identification Model $\lambda_o$}

\begin{figure}[!htb]
\centering
	\includegraphics[width=3.2in]{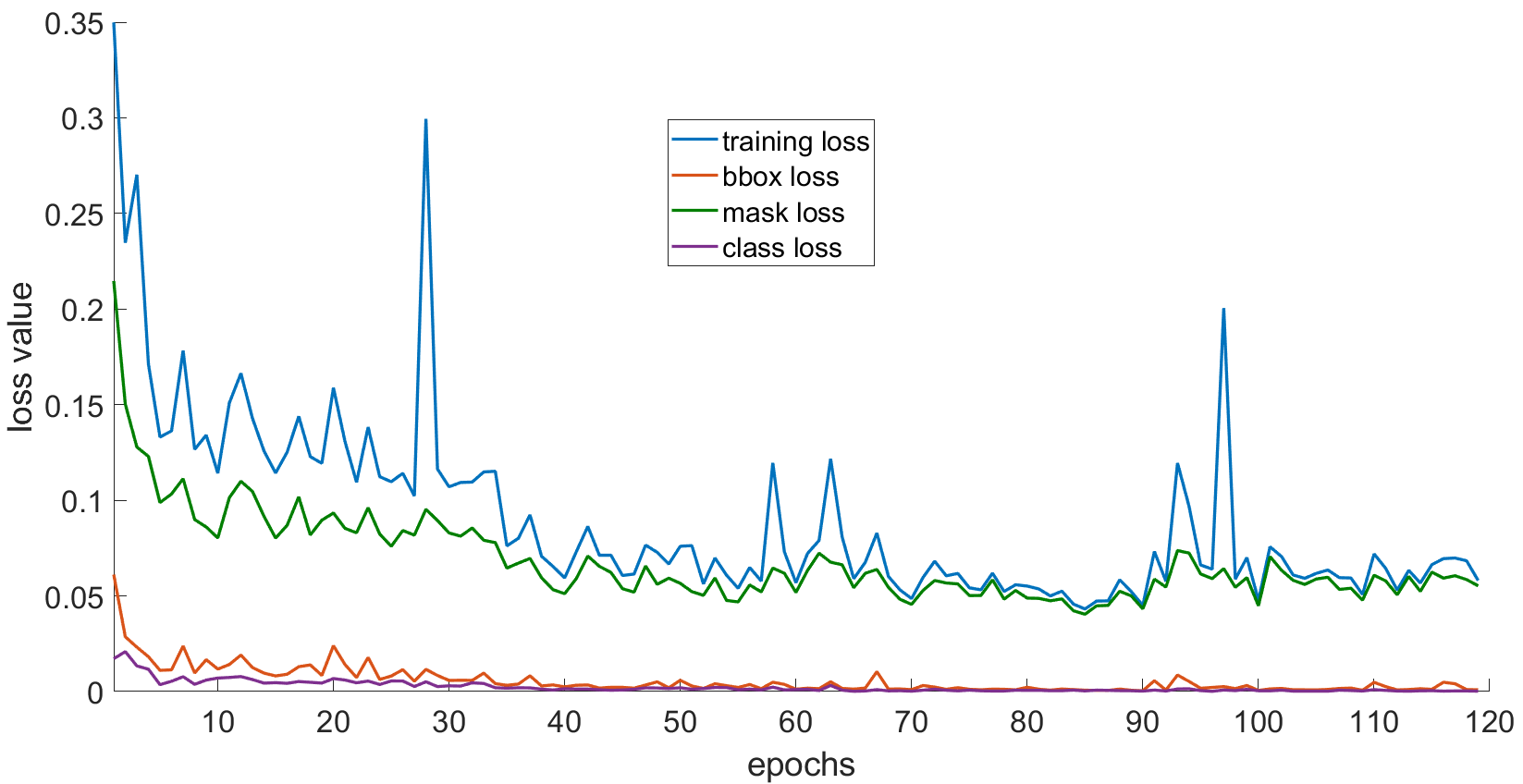}
    \caption{Loss metrics observed for $\lambda_o$}
    \label{fig:mask_rcnn_loss}
\end{figure}

We have used pre-trained weights of COCO model \citep{matterport_maskrcnn_2017} to train $\lambda_o$. The pre-trained network can classify images into 1000 known object categories. To increase the sample size we have used image augmentation such as rotation, flip and skew. Here we also trained multiple networks with various configurations of hyper parameters. Finally, we used $512 \times 512$ as input image dimension and batch size 64. The network is configured as single object classification problem with 2 classes - tower and background. Different loss metrics reported by the network during training phase is shown in Figure \ref{fig:mask_rcnn_loss}. As observed, the trained model converged after around 120 epochs. Thus, we use this model as $\lambda_o$ for tower classification.

\begin{figure}[!htb]
 \centering
    \includegraphics[width=3.2in]{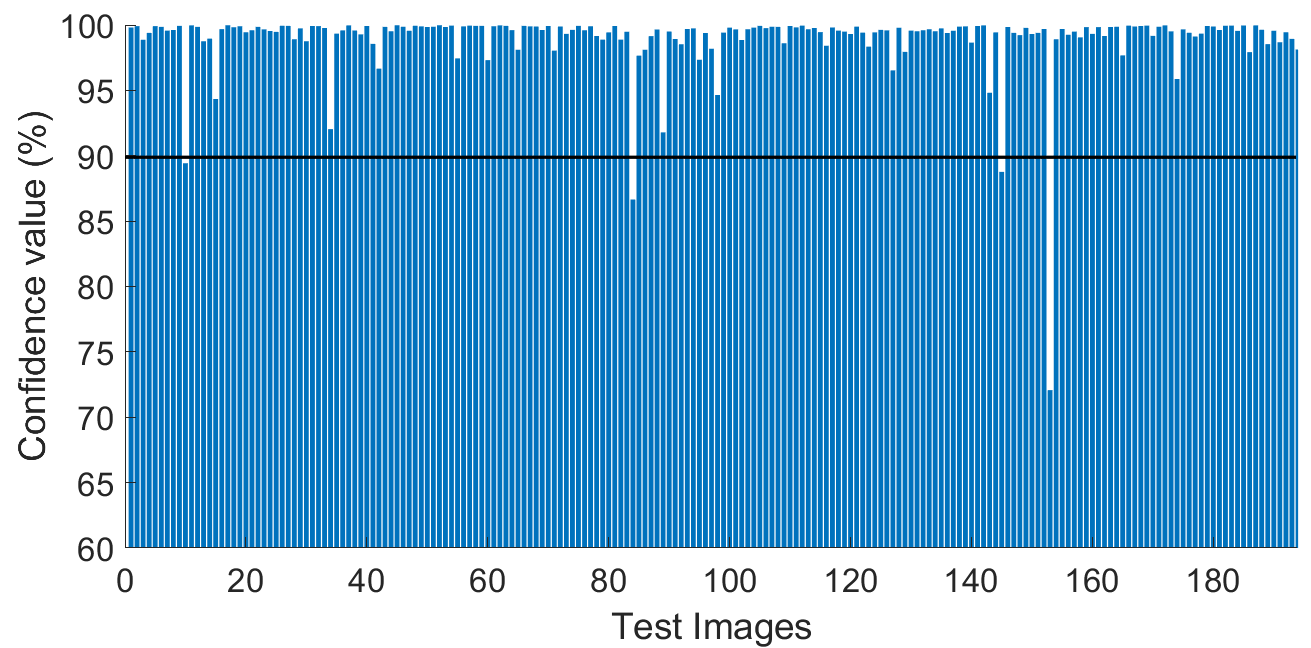}
    \caption{Confidence value of tower object recognition over 194 test images}
    \label{fig:confidence_bar}
\end{figure}

We have used $\tau_o=0.9$, that is, the confidence value over 90\% is considered as correct recognition. Based on this condition in 190 out of 194 images tower is classified with confidence $\geq 0.9$ and while the other 4 were below the threshold (see Figure \ref{fig:confidence_bar}), therefore object recognition accuracy is 98\%.

\begin{figure}[!htb]
\centering
\begin{subfigure}{.5\textwidth}
    \centering
	\includegraphics[width=2.7in]{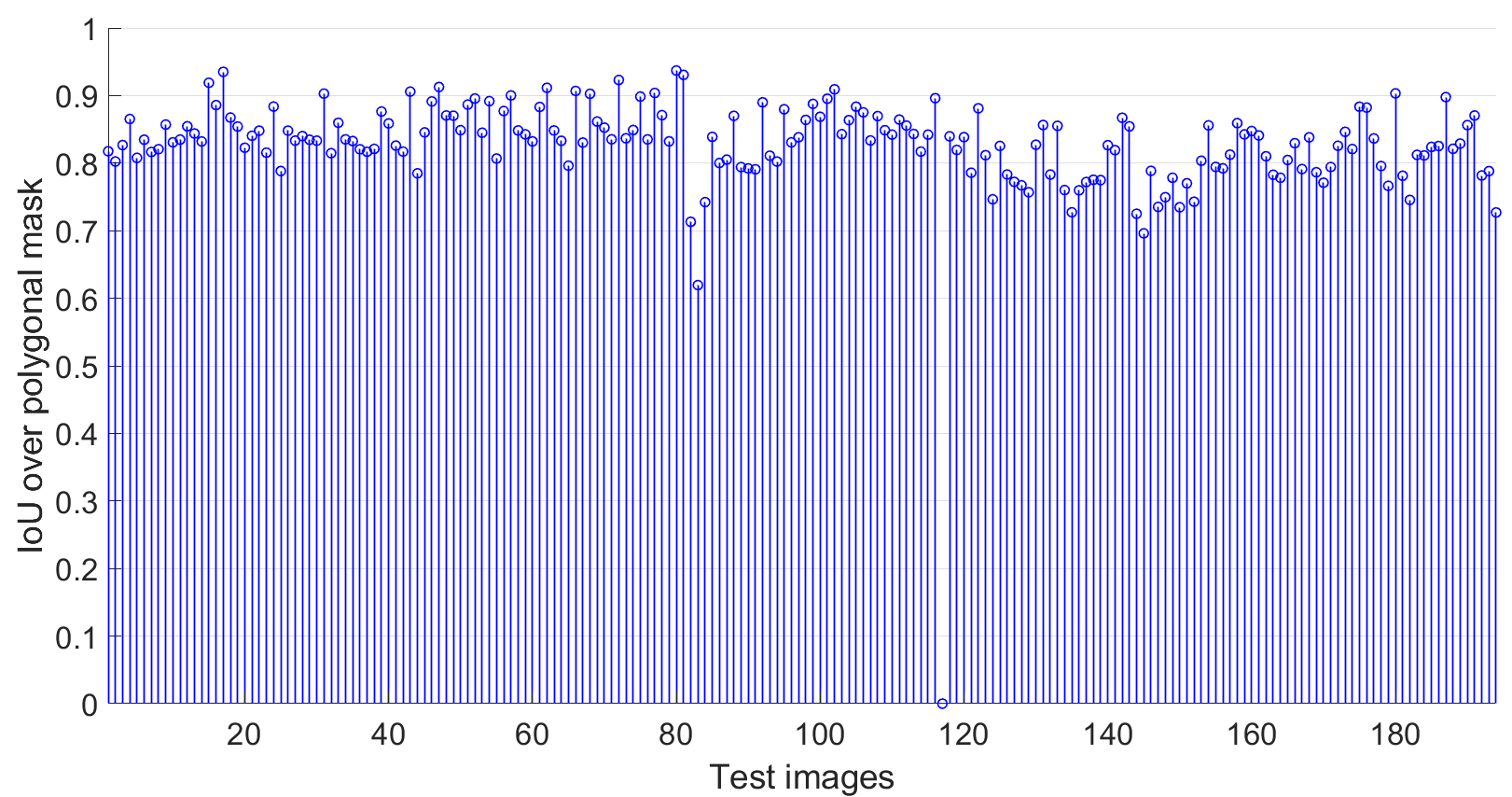}
	\caption{polygonal mask }
	 \label{fig:mask_iou}
\end{subfigure}%
\begin{subfigure}{.5\textwidth}
 \centering
    \includegraphics[width=2.7in]{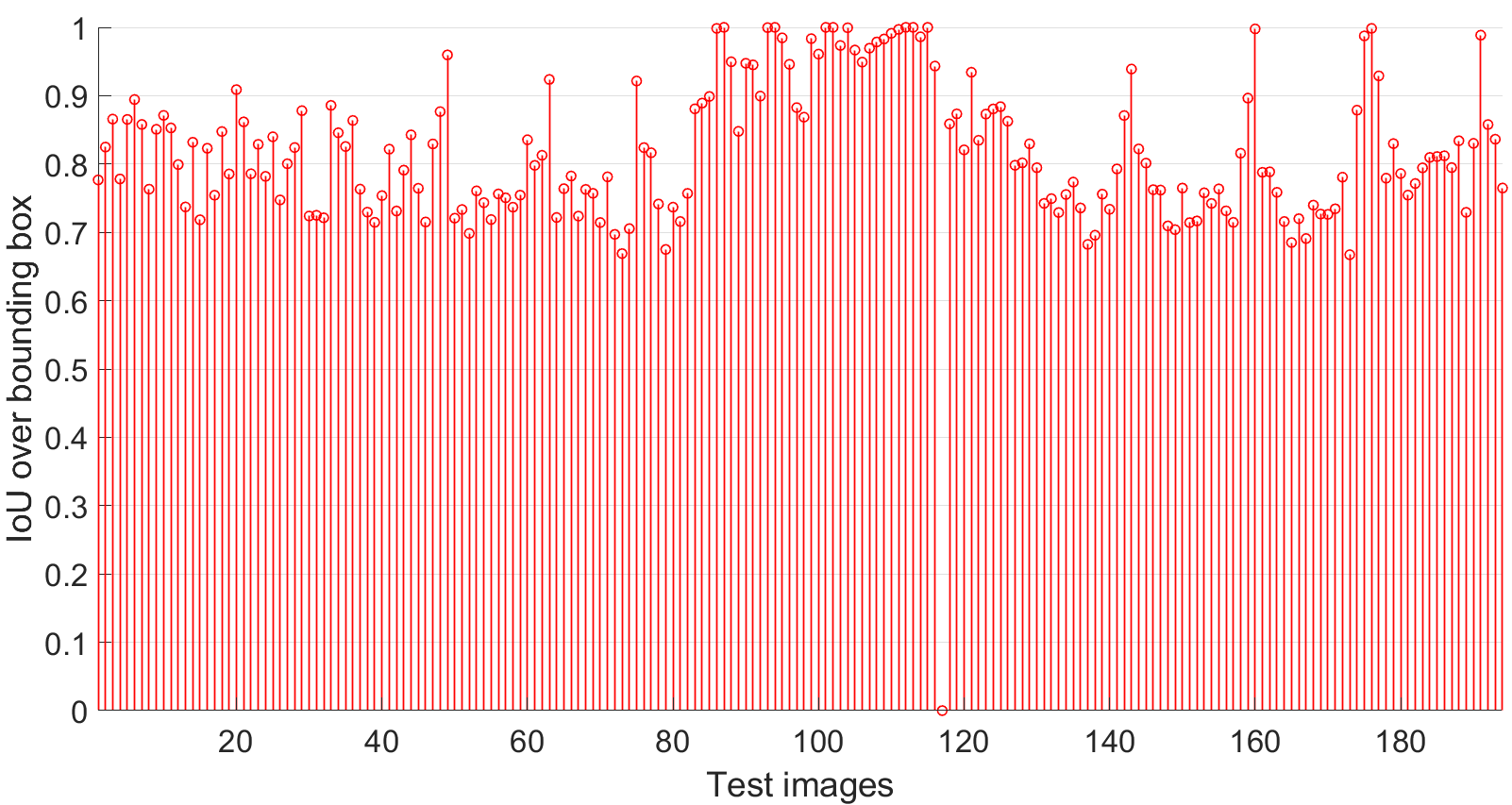}
    \caption{bounding box}
	\label{fig:bbox_iou}
\end{subfigure}
\caption{IoU values}
\end{figure}

   

     

The IoU values for mask and bbox according to Equation \ref{eq:iou} and \ref{eq:iou_bb} are also computed for 194 test images which is presented in Figure \ref{fig:mask_iou} and \ref{fig:bbox_iou}. As observed from IoU values for both processes, the mask for only one image was not predicted correctly. Other images, satisfies the condition for true positive for $IOU \geq 0.5$ for both mask and bounding box overlap. Table \ref{tab:iou} shows the performance of $\lambda_o$ using precision ($IOU_{th}$ values) from 0.5 to 0.75 with a step size of 0.05 \citep{rezatofighi2019generalized} and final AP value. The $th$  value is considered till 0.75 as the precision value significantly deteriorated after that. In both cases $\lambda_o$ showed very high AP value ( 98.11\% for polygonal mask and 94.42\% for bounding box). 


\begin{table}[!htb]
  \caption{Precision and Average Precision (AP) using IoU  over mask and bounding box}
  \label{tab:iou}
  \centering
  \begin{tabular}{|l|l|l|l|l|l|l|l|}
    \hline
    \textbf{IoU} & $IoU_{0.5}$ & $IoU_{0.55}$ & $IoU_{0.6}$ &  $IoU_{0.65}$ & $IoU_{0.7}$ &  $IoU_{0.75}$ &\textbf{ AP} \\
    \hline
   \textbf{ $P_{mask}$} & 99.49\% & 99.49\% & 99.49\% & 98.97\% & 98.45\% &  92.78\% & \textbf{98.11\%} \\
   \hline
    \textbf{ $P_{bbox}$} & 99.49\% & 99.49\% & 99.49\% & 99.49\% & 94.85\% &  73.71\% & \textbf{94.42\%} \\
    \hline
\end{tabular}
\end{table}

\subsubsection{Training and Evaluation for the ensemble model, $\lambda$}
As presented in Table \ref{tab:evaluation} we did not obtain good precision for $\lambda_c$ due to higher false positives (FPs). To improve the precision in $\lambda_c$ (i.e. minimise FPs) we developed the ensemble model, $\lambda$, by combining outcomes of $\lambda_c$ and $\lambda_o$. As described in Algorithm \ref{alg:erc}, two feature sets ($FC$ and $FB$) are generated using binary outcomes 
and confidence values of  $\lambda_c$ and $\lambda_o$. Both $FC$ and $FB$ (presented in Algorithm \ref{alg:erc}) are randomised and fed into the selected machine learning models. The results obtained from both feature sets over the best performing machine learning classifiers are presented in Table \ref{tab:evaluation_ensemble}.

\begin{table}[!htb]
  \caption{Performance of $\lambda$ using $FB$ and $FC$ as feature set and different ML classifier}
  \label{tab:evaluation_ensemble}
  \centering
  \begin{tabular}{|p{1.6cm}|l|l|l|l|l|l|l|l|l|}
    \hline
    {\textbf{Ensemble model}} & \multicolumn{2}{c|} {\textbf{Acc.(\%)}} &  \multicolumn{2}{c|}{\textbf{P(\%)}} & \multicolumn{2}{c|}{\textbf{R(\%)}} & \multicolumn{2}{c|}{\textbf{F1(\%)}} \\
    & \textbf{FB} & \textbf{FC} & \textbf{FB} & \textbf{FC} & \textbf{FB} & \textbf{FC} & \textbf{FB} & \textbf{FC}\\
    \hline
    MLP  & 93.80 & 83.25  & 88.00 & 92.30  & 93.80 & 83.20 & 90.80 & 86.80\\
    \hline
    SVM  & 93.80 & 83.70  & 88.00 & 92.30  & 93.80 & 83.70 & 90.80 & 87.10\\
    \hline
    XGBoost & 93.80 & 86.29  & 88.00 & 92.10  & 93.80 & 86.30 & 90.80 & 88.70\\
    \hline
\end{tabular}
\end{table}

We observed higher precision values as well as accuracy, recall and F1 in $\lambda$ than $\lambda_c$ (which was 53.2\% before) over all 3 classifiers (MLP, SVM and XGBoost).  For, $FB$ all classifiers showed same performance in terms of accuracy, precision, recall and F1-score. However, the performance of XGBoost was the best among all in $FC$ in terms of accuracy and F1-score. In $\lambda$ we observed 8.72\% improvement accross all 3 classifiers in accuracy over the SLP in $\lambda_c$.
The final evaluation outcome from $\lambda$, provides evidence that \model is an effective model for corrosion detection. The ensemble model over multiple CNNs can detect corrosion on real-world images with 93.8\% accuracy and with 88\% precision and 90.8\% F1-score. Our evaluation results show the validity of our primary motivation that utilising an ensemble of CNN models can be effectively used for corrosion detection in drone images of complex industrial structure.

\section{Conclusion}
\label{sec:conclusion}

We have presented \model, an ensemble AI framework underpinned by CNN models for structural identification and corrosion detection in civil engineering settings. Our approach relies on standard approaches to drone managed image capture technologies used in the risk-based maintenance philosophy. This industry conforming approach makes \model applicable in standard working practises where images are used to carry out in-person inspections due to either safety concerns or inaccessibility. Our approach provides engineers with advanced indication of which images may contain sufficient corrosion to warrant maintenance intervention. This in turn provides an opportunity for fine tuning analysis and assessment further along the value chain of the maintenance life cycle by allowing resources to be directed only to those structures highlighted by \model. This opportunity can greatly reduce financial costs and extend the resources of engineers to consider far more images than would otherwise be possible in a human based analysis. 


We evaluated \model via empirical evaluations with state-of-the-art that also utilise AI techniques for structural corrosion analysis.  We demonstrated that our approach is a significant improvement and achieved an overall success rate in excess of 92\%. This level of success indicates a level of efficacy that will allow our approach to provide the foundations for building engineering analysis tools, especially in the area of supporting telecommunication tower safety and maintenance life cycle services. Future work will explore applying \model in a broader range of structures and settings. 



\bibliographystyle{model1-num-names}
\bibliography{corr_detect.bib}







\end{document}